\documentclass[letterpaper, 10 pt, conference, nofonttune]{ieeeconf}
\IEEEoverridecommandlockouts
\title{\LARGE \bf Multi-layered Safety of Redundant Robot Manipulators via Task-oriented Planning and Control
}

\usepackage{cite}
\usepackage{amsmath,amssymb,amsfonts}
\usepackage{algorithmic}
\usepackage{graphicx}
\usepackage{textcomp}
\usepackage{xcolor}
\usepackage{algorithm}
\usepackage{multirow}
\usepackage{amsmath} 
\usepackage{bm}
\usepackage{siunitx}

\usepackage{array}
\usepackage{tabularx}
\usepackage{booktabs}
\usepackage[super]{nth}

\usepackage[caption=false,font=footnotesize]{subfig}

\usepackage{url}
\usepackage[colorlinks,linkcolor=black,anchorcolor=black,citecolor=black,urlcolor=black]{hyperref}

\usepackage{threeparttable}

\usepackage{microtype} 

\let\labelindent\relax
\usepackage{enumitem}

\author{Xinyu Jia, Wenxin Wang, Jun Yang, Yongping Pan, \IEEEmembership{Senior Member, IEEE}, \\
and Haoyong Yu, \IEEEmembership{Senior Member, IEEE}% <-this % stops a space
\thanks{This work was supported in part by the Science and Engineering Research Council, Agency of Science, Technology and Research, Singapore, through the National Robotics Program under Grant No. M22NBK0108, and in part by the Major Key Project of PCL, China under Grant No. PCL2024A04.}% <-this % stops a space
\thanks{Xinyu Jia, Wenxin Wang, Jun Yang, and Haoyong Yu are with the Department of Biomedical Engineering, National University of Singapore, 117583, Singapore. (Corresponding author: Haoyong Yu. Email: {\tt\small bieyhy@nus.edu.sg})}%
\thanks{Yongping Pan is with the Peng Cheng Laboratory, Shenzhen 518057, China, and also the School of Electrical and Electronic Engineering, Nanyang Technological University, 639798, Singapore.}
}

\begin{document}

\maketitle
\thispagestyle{empty}
\pagestyle{empty}

\begin{abstract}
    Ensuring safety is crucial to promote the application of robot manipulators in open workspaces. Factors such as sensor errors or unpredictable collisions make the environment full of uncertainties. In this work, we investigate these potential safety challenges on redundant robot manipulators, and propose a task-oriented planning and control framework to achieve multi-layered safety while maintaining efficient task execution. Our approach consists of two main parts: a task-oriented trajectory planner based on multiple-shooting model predictive control (MPC) method, and a torque controller that allows safe and efficient collision reaction using only proprioceptive data. Through extensive simulations and real-hardware experiments, we demonstrate that the proposed framework\footnote{Code is available at \href{https://github.com/jia-xinyu/arm-safety}{https://github.com/jia-xinyu/arm-safety}.} can effectively handle uncertain static or dynamic obstacles, and perform disturbance resistance in manipulation tasks when unforeseen contacts occur.
\end{abstract}

% \begin{IEEEkeywords}
% Collision reaction, estimated torque, physical human-robot interaction (pHRI), robotic manipulator
% \end{IEEEkeywords}

\section{Introduction}

\begin{figure}[!t]
    \centering
    \includegraphics[width=0.38\textwidth]{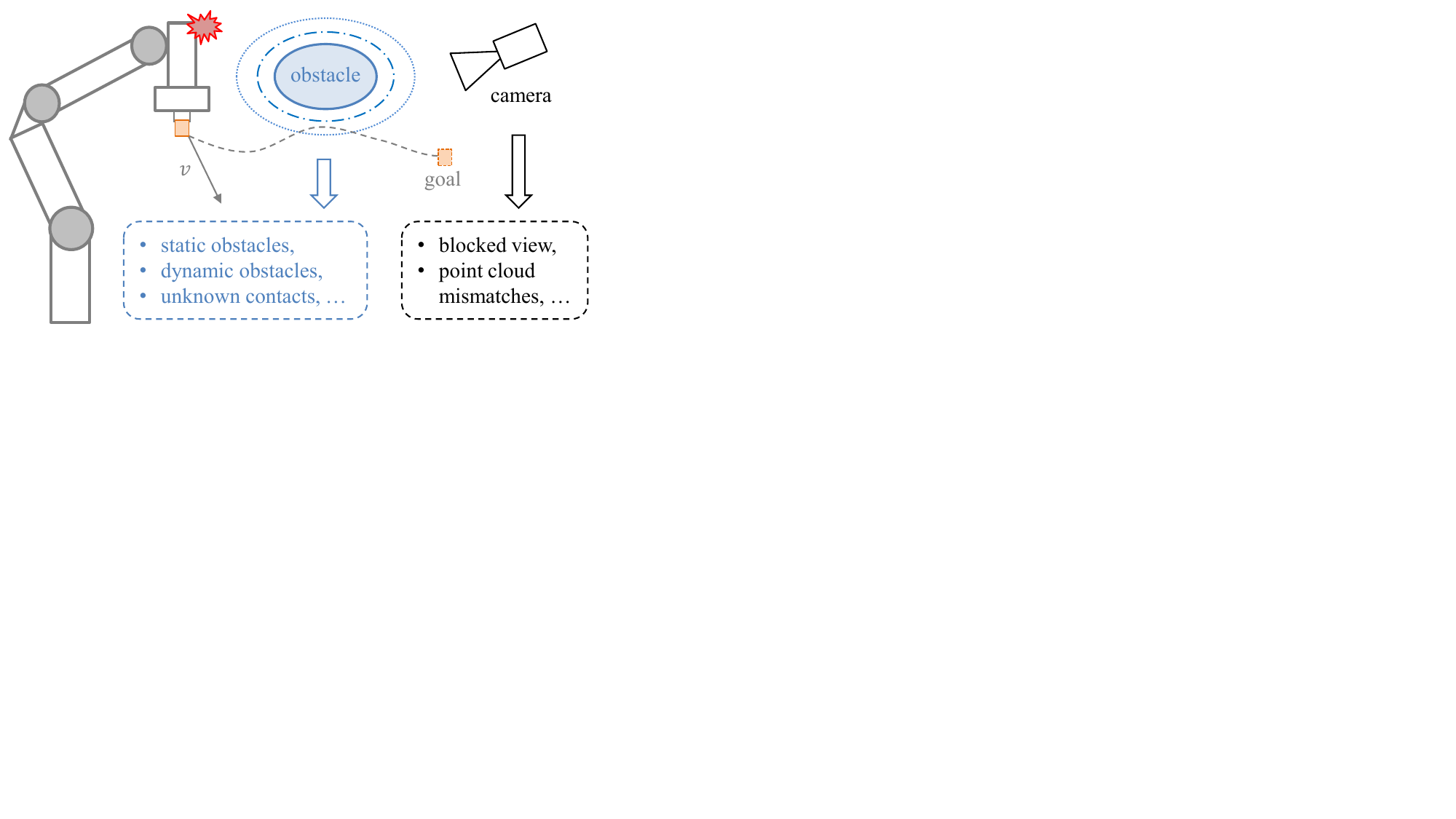}
    \caption[]{When a robot operates in an open workspace, static and dynamic obstacles, as well as unknown contacts, need to be considered to ensure safety, while minimizing their impact on task execution is desirable.}
    \label{fig:cover}
    \vspace{-0.3cm}
\end{figure}

The safety of robot manipulators in open workspaces is a major concern in the robotics community \cite{aude_review_2019}. Unlike industrial assembly lines with isolated areas, open workspaces bring higher environmental uncertainties \cite{Sami_2023}. For example, cameras may fail to accurately model the environment due to occlusions, or external contacts occur without prediction (see Fig. \ref{fig:cover}). Such unforeseen events can cause damage to robots and threaten nearby humans. Fenceless robots typically define different safety zones to regulate their movements \cite{mcri_collaborative_2015}. However, these zones are often coarsely defined, and manipulation is interrupted whenever a human is detected entering them. In this paper, we consider the scenarios where these safety challenges arise simultaneously and aim to balance safety and task execution when the two objectives conflict.

In obstructed environments, a common solution is to generate collision-free trajectories through motion planning based on sensing information. Well-known approaches include sampling-based methods, such as the probabilistic roadmap (PRM) \cite{Kavraki_prm_1996} and the rapidly-exploring random tree (RRT) \cite{LaValle_rrt_1998}. They can plan global trajectories offline in high-dimensional spaces but struggle in multi-objective and multi-constraint scenarios. Optimization-based methods, such as the covariant Hamiltonian optimization for motion planning (CHOMP) \cite{Ratliff_chomp_2009} or the stochastic trajectory optimization for motion planning (STOMP) \cite{Kalakrishnan_stomp_2011}, can produce locally optimal solutions incorporating trajectory smoothness into obstacle avoidance. However, they are typically limited to optimizing in joint space without considering tasks in Cartesian space. In contrast, inverse kinematics (IK) solvers, such as the RelaxedIK \cite{Rakita_released_ik_2018}, leverage the kinematic redundancy, allowing the robot to avoid obstacles while still preserving manipulation. Unfortunately, these solvers only focus on instantaneous control and hence lack prediction capabilities. Recently, control barrier function (CBF) schemes have received much attention \cite{Singletary_cbf_2022, Jang_cbf_2024}, which restrict control inputs to ensure safety but face similar limitations as IK solvers. In comparison, model predictive control (MPC) combines offline planning with online execution, allowing for future state prediction as well as incorporating feedback signals \cite{Lee_mpc_2023}. While many studies explore its application for obstacle avoidance \cite{eth_mpc_2020, eth_mpc_2022}, few take into account the trade-off between safety and task execution.

When the environment cannot be precisely interpreted, system safety is compromised. In fact, obtaining absolutely accurate geometric information in the real world is challenging. Therefore, it is essential to develop safety strategies to handle unforeseen collisions. Typically, collision detection relies on generalized momentum observers \cite{sami_collision_2017, Jia_icra_2024}. Once a potential collision is identified, the simplest and most common response is to immediately brake the actuators \cite{sami_franka_2022}. However, terminating manipulation is not always accepted because it can reduce productivity. Moreover, locking the joints may lead to additional collisions if the object in contact is still moving. \cite{Luca_contact_2008} provides an idea that exploits the null space of redundant robots to preserve Cartesian task execution despite the occurrence of collisions. In contrast, contact-driven approaches offer more aggressive strategies, allowing the posture task to continue along the contact constrained direction \cite{Jorda_contact_2019}. Nevertheless, these methods may cause false triggers or repeated collisions in dynamic environments. A conservative reaction strategy might be appropriate, particularly in human-robot interactions \cite{Sami_2023}.

In this paper, the safety of redundant robot manipulators in complex operational environments is systematically investigated. Specifically, the safety is divided into the following three layers to achieve: (\textbf{S1}) when approaching to a known obstacle, the nearest robot links move away from the obstacle without affecting task execution, so as to leave sufficient clearance between the robot and the obstacle for low collision risks; (\textbf{S2}) safety is always the top priority, so the robot bodies never come into contact with the known obstacle, even though the manipulation task has to be affected; (\textbf{S3}) the robot complies with unforeseen contacts in the environment, while performing disturbance rejection on manipulation tasks.

\begin{figure}[!t]
    \centering
    \includegraphics[width=0.48\textwidth]{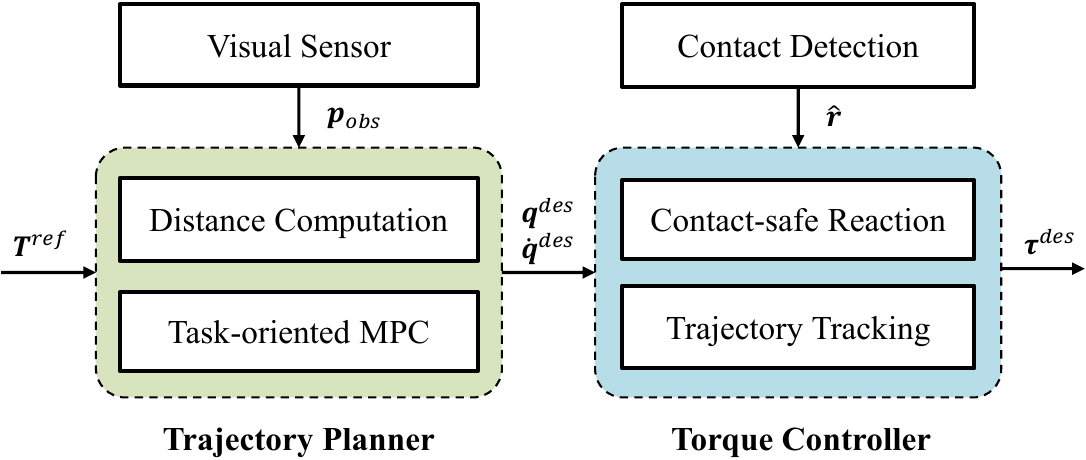}
    \caption[]{A schematic diagram depicting the various components of the proposed framework. The high-level planning module is to avoid known obstacles, while the cascaded control module aims to handle unknown contacts. }
    \label{fig:framework}
    \vspace{-0.3cm}
\end{figure}

To fulfill these requirements, we propose a task-oriented planning and control framework, as shown in Fig. \ref{fig:framework}. This framework leverages multi-objective optimization and redundancy advantages to ensure hierarchical safety of the whole robot body. The main contributions of this work are as follows:
\begin{enumerate}[label=\textbullet, left=1em]
    \item We propose a novel trajectory planner based on nonlinear MPC to address \textbf{S1} and \textbf{S2}.  In particular, to enable fast computation for long-horizon planning, we incorporate multiple shooting and local linearization techniques.
    \item A control strategy that switches between trajectory tracking and contact-safe modes is developed to generate safe and efficient reaction behaviors in the \textbf{S3} scenario. 
    \item We validate the proposed framework through extensive simulations and real hardware experiments on a redundant manipulator in three case studies, including static obstacles, dynamic obstacles and unknown contacts. 
\end{enumerate}

\section{Modeling of Redundant Manipulators}
\label{system}

A redundant manipulator with $n$ degrees of freedom (DoFs) in $m \left(m < n \right)$ dimensional task space satisfies $\bm{T} = \text{FK}(\bm{q})$, where $\bm{T} \in SE(3)$ is the end-effector pose, $\bm{q} \in \mathbb{R}^n$ is the joint position, $\text{FK}(\cdot)$ is the forward kinematics. According to \cite{modern_robotics_2017}, by taking the time derivative of kinematics, the mapping between task space and joint space can be expressed as
\begin{equation}
    \label{eq:fw_kin}
    \bm{\mathcal{V}} = \bm{J}(\bm{q})\dot{\bm{q}},
\end{equation}

\noindent where $\bm{J} \in \mathbb{R}^{m\times n}$ is the Jacobian matrix, $\bm{\mathcal{V}} \in \mathbb{R}^m$ is the end-effector twist (or called spatial velocity). If the full Cartesian space ($m = 6$) is considered, the twist will contain three translational velocities and three rotational velocities. Since the considered system is kinematically redundant, more than one configuration exists for a given end-effector pose, which allows the robot to perform multiple tasks simultaneously. The additional null space of dimension $n - m$ is defined as
\begin{equation}
    \label{eq:inv_kin}
    \bm{N}(\bm{q}) = \bm{I} - \bar{\bm{J}}\bm{J}, 
\end{equation}

\noindent where $\bar{\bm{J}} = \bm{J}^\top(\bm{J} \bm{J}^\top)^{-1}$ is the pseudo inverse of Jacobian, $\bm{N}$ is the corresponding null space projector. 

Furthermore, the robot dynamics can be expressed as
\begin{equation}
    \label{eq:dyn_joint}
    \bm{M}(\bm{q})\ddot{\bm{q}} + \bm{C}(\bm{q},\dot{\bm{q}})\dot{\bm{q}} + \bm{g}(\bm{q}) = \bm{\tau},
\end{equation}

\noindent where $\bm{M}\in \mathbb{R}^{n\times n}$ the inertia matrix, $\bm{C}\in \mathbb{R}^{n\times n}$ is the Coriolis and centrifugal matrix, $\bm{g}\in \mathbb{R}^n$ is the gravity vector, $\bm{\tau} \in \mathbb{R}^n$ is the joint torque. The task-space dynamics can be obtained by multiplying Eq. (\ref{eq:dyn_joint}) by $\bar{\bm{J}}^\top$:
\begin{equation}
    \label{eq:dyn_task}
    \bm{\Lambda}(\bm{q})\dot{\bm{\mathcal{V}}} + \bm{\eta}(\bm{q},\dot{\bm{q}}) = \bm{\mathcal{F}},
\end{equation}

\noindent where $\bm{\Lambda} = \bar{\bm{J}}^\top\bm{M}\bar{\bm{J}}$ is the task inertia matrix, $\bm{\eta} = \bar{\bm{J}}^\top (\bm{C}\dot{\bm{q}} + \bm{g}) - \bm{\Lambda}\dot{\bm{J}}\dot{\bm{q}}$ denotes the Coriolis and gravity terms projected in the task space, $\bm{\mathcal{F}}$ is the end-effector wrench (or called spatial force). In some literature \cite{kim_highly_2019}, $\bar{\bm{J}}$ is replaced by the dynamically consistent inverse $\bm{M}^{-1}\bm{J}^\top\bm{\Lambda}$ for computing Eq. (\ref{eq:dyn_task}). However, as the acceleration does not participate in prioritizing tasks in this paper, the SVD-based pseudo inverse defined in Eq. (\ref{eq:inv_kin}) is still preserved. Moreover, for the sake of distinction, we use $\bm{J}_t$ to denote the generic task Jacobian ($m \leq 6$) in the following while $\bm{J}$ is only expressed in the Cartesian space.

\section{Task-oriented planning}
\label{planning}
In this section, we present a task-oriented trajectory planner based on nonlinear MPC to achieve \textbf{S1} and \textbf{S2}. The planner aims to predict whole-body motion in real time according to current obstacle distances and generate feasible trajectories with minimal task execution cost. For this multi-objective optimization problem, we set $\bm{x} = (\bm{q}, \dot{\bm{q}}) \in \mathbb{R}^{2n}$, $\bm{u} = \ddot{\bm{q}} \in \mathbb{R}^{n}$ as the state and input variables. Their discrete forms are
\begin{equation}
    \label{eq:discrete_kin}
    \left\{
    \begin{aligned}
    &\bm{q}_{k+1} = \bm{q}_k + \Delta t \cdot \dot{\bm{q}}_k, \\
    &\dot{\bm{q}}_{k+1} = \dot{\bm{q}}_k + \Delta t \cdot \ddot{\bm{q}}_k,
    \end{aligned}
    \right.
\end{equation}

\noindent where $\Delta t$ is the time step closely related to the prediction horizon. Then, a classical MPC formulation is given as
\begin{align}
\label{eq:mpc}
\left[\mathbf{X}^*, \mathbf{U}^*\right] &= \arg\min_{\mathbf{X}, \mathbf{U}} L_f(\bm{x}_{N}) + \sum_{k=0}^{N-1} L(\bm{x}_k, \bm{u}_k) \notag \\
\text{s.t.} \quad 
&\bm{x}_{k+1} = \bm{f}(\bm{x}_k, \bm{u}_k), \quad \bm{x}_0 = \bm{x}(0), \\
&\bm{x}_k \in \mathcal{X}, \quad \forall k=\{0, 1, \dots, N\}, \notag \\
&\bm{u}_k \in \mathcal{U}, \quad \forall k=\{0, 1, \dots, N-1\}, \notag
\end{align}

\noindent where $\mathbf{X} \!=\! \{\bm{x}_0, \bm{x}_1, \cdots, \bm{x}_N\}$ and $\mathbf{U} \!=\! \{\bm{u}_0, \bm{u}_1, \cdots, \bm{u}_{N-1}\}$ represent the state and input variable sequences, the superscript $(\cdot)^{*}$ denotes the optimized solution, $\mathcal{X}$ and $\mathcal{U}$ are their respective constraint sets, $N$ is the prediction horizon length, $L(\bm{x}_k, \bm{u}_k)$ and $L_f(\bm{x}_N) $ are the stage cost and terminal cost, $\bm{f}(\bm{x}_k, \bm{u}_k)$ is the state-space model derived from Eq. (\ref{eq:discrete_kin}), and the current measurement $\bm{x}$ is assigned to the initial state $\bm{x}_0$. When the MPC is solved iteratively, the first element of the predicted trajectory, i.e., $\bm{x}_1$, is extracted and tracked by the cascaded torque controller. In the following, we will describe the design of cost terms and constraints in detail.

\subsection{Cost Function}
    The main goal of the cost function is to accomplish tasks in the Cartesian space. To this end, the stage cost is designed as 
    \begin{equation}
        \label{eq:cost_function}
        L(\bm{x}_k, \bm{u}_k) = L_{ee}(\bm{x}_k) + L_{rep}(\bm{x}_k) + L_{s}(\bm{x}_k) + \| \bm{u}_k \|^2_{\bm{R}}, 
    \end{equation}
    \noindent with
    \begin{equation}
        \label{eq:cost_error}
        L_{ee}(\bm{x}_k) = \| \bm{\mathcal{V}}^{ref} - \bm{\mathcal{V}}_k  \|^2_{\bm{Q}_{ee}}, \
        L_{s}(\bm{x}_k) = \| \dot{\bm{q}}_k \|^2_{\bm{Q}_{s}},
    \end{equation}
    
    \noindent where $L_{ee}(\bm{x}_k)$ regulates the end-effector to the reference pose, $L_{rep}(\bm{x}_k)$ is a soft constraint associated with obstacle avoidance and will be discussed in Section \ref{task-oriented}, $L_{s}(\bm{x}_k)$ regulates the joint velocity for trajectory smoothness, $\| \bm{u}_k \|^2_{\bm{R}}$ is the penalization on control input, $\bm{Q}_{ee}$, $\bm{Q}_{s}$, $\bm{R}$ are positive diagonal weighting matrices. Note that we utilize a linear twist error $\bm{\mathcal{V}}^{ref} - \bm{\mathcal{V}}_k$ rather than a nonlinear pose error $\bm{T}(\bm{q}_k) \ominus \bm{T}^{ref}$ to represent the end-effector's motion. This approach makes it easier for nonlinear programming (NLP) solvers to find a feasible solution. Here, $\ominus$ denotes the difference in $SE(3)$ and converts to $\mathbb{R}^{6}$, i.e., $\bm{T}_1 \ominus \bm{T}_2 = \log(\bm{T}^{-1}_1 \bm{T}_2)$, the superscript $( \cdot )^{ref}$ indicates the reference, and the two twists in Eq. (\ref{eq:cost_error}) can be computed by
    \begin{equation}
        \label{eq:cost_twist}
        \bm{\mathcal{V}}^{ref} = \bm{T}(\bm{q}) \ominus \bm{T}^{ref}, \
        \bm{\mathcal{V}}_k = \bm{J}(\bm{q})(\bm{q}_k - \bm{q}),
    \end{equation}
    
    \noindent where $\bm{T}$ and $\bm{J}$ are obtained from the current measurement $\bm{q}$. For the terminal cost, $L_f(\bm{x}_N) = L_{ee}(\bm{x}_N) + L_{s}(\bm{x}_N)$.

\subsection{Multiple Shooting}
    To improve numerical behavior of the MPC, the multiple shooting method is employed by introducing $\bm{x}_k$ as additional decision variables (i.e., shooting nodes) and setting equality constraints at each node as follows 
    \begin{equation}
        \label{eq:shooting}
        \bm{g}_{k+1} = \bm{x}_{k+1} - \bm{f}(\bm{x}_k, \bm{u}_k) = \bm{0},
    \end{equation}
    
    \noindent where $\bm{g}_{k+1}$ represents the gap between the next shooting state $\bm{x}_{k+1}$ and the predicted state $\bm{f}(\bm{x}_k, \bm{u}_k)$ at $k + 1$.	Although this method adds intermediate variables and new constraints, it effectively mitigates the nonlinearity propagation in single shooting schemes \cite{Carlos_ddp_2023}. As a result, it will lead to faster convergence and better numerical stability, particularly for trajectory planning scenarios with long prediction horizons.

    % $\bm{g}_1 = \bm{x}_0 - \bm{x}$
 
\subsection{Collision Avoidance}
\label{avoidance}
    For collision avoidance using known environmental information, we represent the robot and obstacles as primitive bodies and employ the algorithm outlined in \cite{Pan_fcl_2012} to query the minimum distance between each pair of bodies. The distance between the robot link $i$ and the obstacle is expressed as
    \begin{equation}
        \label{eq:distance}
        d_i(\bm{q}) = \| \bm{p}_{i_A}(\bm{q}) - \bm{p}_{i_B}(\bm{q}) \|,
    \end{equation}

    \noindent where $d_i$ denotes the positive distance function in collision-free cases, $\bm{p}_{i_A}$ and $\bm{p}_{i_B}$ represent the nearest points on the robot and obstacle, respectively. According to \cite{Schulman_loca_2013}, the gradient of the distance function with respect to the joint position is 
    \begin{equation}
        \label{eq:dist_jacobian}
        \frac{\delta d_i}{\delta \bm{q}} = \bm{n}^{\top}_i (\bm{J}_{i_A}(\bm{q}) - \bm{J}_{i_B}(\bm{q})), 
    \end{equation}

    \noindent where $\bm{J}_{i_A}, \bm{J}_{i_B} \in \mathbb{R}^{3\times n} $ are the Jacobians of $\bm{p}_{i_A}, \bm{p}_{i_B}$ in the world frame, $\bm{n}_i \in \mathbb{R}^3$ is the norm vector, i.e., $\bm{n}_i = (\bm{p}_{i_A} - \bm{p}_{i_B}) / \| \bm{p}_{i_A} - \bm{p}_{i_B} \|$.
    Thus, the distance at the $k$th moment can be derived by the following approximation:
    \begin{equation}
        \label{eq:dist_final}
        d_i(\bm{q}_k) \approx d_i(\bm{q}) + \bm{n}^{\top}_i \bm{J}_{i_A}(\bm{q}) (\bm{q}_k - \bm{q}).
    \end{equation}

    \noindent With this setup, the robot is able to avoid obstacles by enforcing inequality constraints $d_i(\bm{q}_k) \ge d_{th_1}$ on each link, where $d_{th_1}$ is the minimum allowable distance threshold.
            
    \begin{figure}[!t]
        \vspace{-0.3cm}
        \centering
        \subfloat[]{\includegraphics[width=.24\textwidth]{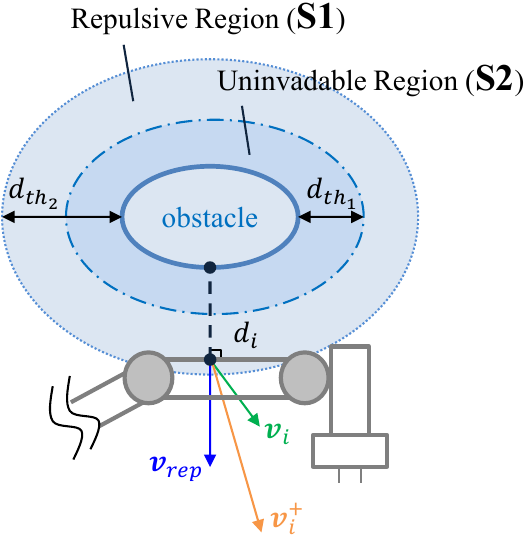}}
        \hfill
        \subfloat[]{\includegraphics[width=.24\textwidth]{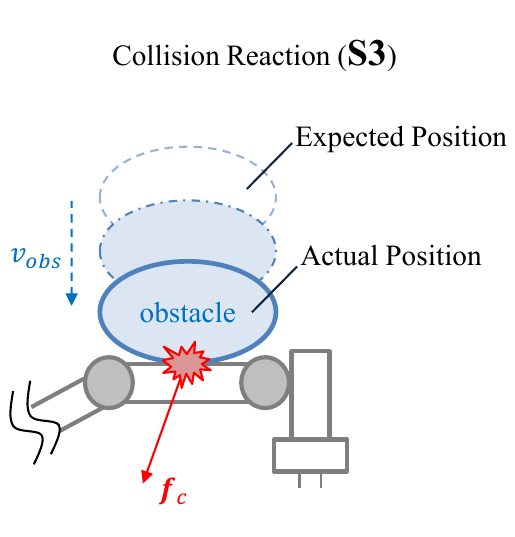}} \\
        \caption[]{Illustration of multi-layered safety in robot manipulation. \textbf{S1} and \textbf{S2} are realized in the trajectory planner. \textbf{S3} is handled by the torque controller.}
        \label{fig:principle}
        \vspace{-0.3cm}
    \end{figure}
    
\subsection{Task-oriented Avoidance}
\label{task-oriented}
    Maintaining the end-effector task while moving links away from obstacles is sometimes not possible since both objectives conflict. Nevertheless, we can exploit kinematic redundancy to harmonize them. Inspired by \cite{Kanazawa_avoid_2021}, a repulsive region (\textbf{S1}) with a distance threshold $d_{th_2}$($d_{th_2} > d_{th_1}$) is defined, as well as an uninvadable region (\textbf{S2}) described in Section \ref{avoidance}, see Fig. \ref{fig:principle}. When the link $i$ enters into the former region, a repulsive velocity $\bm{v}_{rep}\in \mathbb{R}^3$ is generated along $\bm{n}_i$ at the closest point:
    \begin{equation}
        \label{eq:v_rep}
        \bm{v}_{rep} = \bm{n}_i k_{rep} (d_{{th}_2} - d_i),
    \end{equation}

    % \in \mathbb{R}^{+}
    \noindent where $k_{rep}$ is a positive scaling factor. It can be seen that as the link penetrates deeper, the magnitude of $\bm{v}_{rep}$ increases accordingly and reaches the maximum at the boundary of next region. The resulting total velocity $\bm{v}^+_i$ is given by
    \begin{equation}
        \label{eq:v_total}
        \bm{v}^+_i = \bm{v}_i + \bm{v}_{rep},
    \end{equation}

    \noindent where $\bm{v}_i$ denotes the current velocity. This repulsive effect is then incorporated as soft constraints in the cost function as
    \begin{equation}
        \label{eq:cost_rep}
        L_{rep}(\bm{x}_k) = \| \bm{N}_t (\dot{\bm{q}}_k -\bar{\bm{J}}_{i_A} \bm{v}^+_i) \|^2_{\bm{Q}_{rep}},
    \end{equation}
    
    \noindent where $\bm{Q}_{rep}$ is a positive diagonal weighting matrix, $\bm{N}_t$ is the null space 
    of task Jacobian as defined in Eq. (\ref{eq:inv_kin}).
    % $\bm{N}_t = \bm{I} - \bar{\bm{J}}_t\bm{J}_t
    
    % that avoids interfering with the end-effector task. 
    
    Additionally, we consider a goal relaxation term $\lambda \in \left[0, 1\right]$ that acts in the end-effector pose:
    \begin{equation}
        \label{eq:lambda}
        \lambda(d) =
        \begin{cases}
            \exp( - \frac{\alpha(d_{{th}_2} - d)} {d_{{th}_2} -  d_{{th}_1}}), & \text{if } d < d_{{th}_2} \\
            1, & \text{otherwise }
        \end{cases}
    \end{equation}

    \noindent where $\alpha$ a positive shaping factor. With this function, the cost $L_{ee}(\bm{x}_k)$ in Eq. (\ref{eq:cost_error}) is reformulated as
    \begin{equation}
        \label{eq:cost_new}
        L_{ee}(\bm{x}_k) = \| \bm{\mathcal{V}}^{ref} - \bm{\mathcal{V}}_k \|^2_{\lambda(d_i) \bm{S} \bm{Q}_{ee}},
    \end{equation}
    
    \noindent where $\bm{S}$ is a selection matrix isolating all elements in $\bm{Q}_{ee}$ that are irrelevant to end-effector manipulation. The revised pose regulation cost in Eq. (\ref{eq:cost_new}) will decrease the contribution of the goal attraction when the robot approaches the obstacle. 
    
% \subsection{Complementary Remarks}

\section{Control with collision reaction}
\label{control}
The optimal trajectory from the MPC is tracked by high-frequency feedback controllers that compute torque commands based on the dynamics. Notably, the controller can safely react to unpredicted contacts (\textbf{S3}) while keeping task consistency.

\subsection{Trajectory Tracking}
\label{track}
    Since the solutions $\bm{q}^{des}$, $ \dot{\bm{q}}^{des}$ of Eq. (\ref{eq:mpc}) already satisfy a set of prioritized tasks, they are directly tracked in free space by a standard computed torque controller  \cite{modern_robotics_2017} as follows
    \begin{equation}
        \label{eq:tau_ff}
        \bm{\tau}_{ff} = \bm{M} \, (\bm{K}_{p_1}(\bm{q}^{des} - \bm{q}) + \bm{K}_{d_1}(\dot{\bm{q}}^{des} - \dot{\bm{q}}))
        +\bm{C}\dot{\bm{q}} + \bm{g},
    \end{equation}

    \noindent where $\bm{\tau}_{ff}$ is the feedforward torque, $\bm{M}$, $\bm{C}$, $\bm{g}$ are given in Eq. (\ref{eq:dyn_joint}), $\bm{K}_{p_1}$, $\bm{K}_{d_1}$ are diagonal positive matrices of control gain. Practically, another proportional-derivative (PD) portion is integrated to enhance tracking precision, expressed as
    \begin{equation}
        \label{eq:tau_des_free}
        \bm{\tau}^{des} = \bm{\tau}_{ff} + \bm{K}_{p_2}(\bm{q}^{des} - \bm{q}) + \bm{K}_{d_2}(\dot{\bm{q}}^{des} - \dot{\bm{q}}),
    \end{equation}

    \noindent where $\bm{\tau}^{des}$ is the final torque command sent to actuators. It needs to be emphasized that the gains $\bm{K}_{p_2}$ and $\bm{K}_{d_2}$ make a small contribution to the total torque. The actuator behavior is still primarily decided by the feedforward term $\bm{\tau}_{ff}$.
    
\subsection{Contact Detection}
\label{contact_detect}
    When the robot unexpectedly collides with the obstacle or encounters unknown contacts, the dynamics Eq. (\ref{eq:dyn_joint}) becomes 
    \begin{equation}
        \label{eq:dyn_joint_contact}
        \bm{M}\ddot{\bm{q}} + \bm{C}\dot{\bm{q}} + \bm{g} = \bm{\tau} + \bm{\tau}_c,
    \end{equation}
    
    \noindent where $\bm{\tau}_c$ is the external torque. Here, we leverage the unknown system dynamics estimator (USDE) in our previous work \cite{Jia_tmech_2024} to obtain the estimation of external torque 
    \begin{equation}
        \label{eq:usde}
        \hat{\bm{r}} = \frac{\bm{\mathcal{P}} - \bm{\mathcal{P}}_f} {k} + \bm{\mathcal{H}}_f - \bm{\tau}_f,
    \end{equation}
    \noindent with 
    \vspace{-0.2cm}
    \begin{equation}
        \label{eq:auxiliary}
        \bm{\mathcal{P}} = \bm{M}\dot{\bm{q}},  \quad 
        \bm{\mathcal{H}} = - \bm{C}^\top \dot{\bm{q}} + \bm{g},
    \end{equation}
    
    \noindent where $\bm{\mathcal{P}}$ and $\bm{\mathcal{H}}$ serve as auxiliary variables, the subscript $(\cdot)_f$ indicates the first-order filtered form of associated variables, $k$ is a positive filter coefficient, and $\hat{\bm{r}}$ is the estimated result. As proven in \cite{Jia_tmech_2024, yang_unknown_2020}, $\bm{\tau}_c \approx \hat{\bm{r}}$ when $k\to0$. In practice, the collision is determined if the element of $\hat{\bm{r}}$ exceeds a predefined threshold $\tau_{th}$. The collision link is identified according to the element of $\hat{\bm{r}}$ that is above $\tau_{th}$ and furthest from the robot base.

\subsection{Contact-safe Reaction}
\label{contact_safe}
    As shown in Fig. \ref{fig:principle} (b), once the robot detects a collision, it should respond to ensure safety. Rather than relying on simple shutdown protection, we aim to continue the manipulation with minimal effect from the collision. Similar to Section \ref{task-oriented}, the redundancy is leveraged to coordinate the two objects. First, the dynamics in the end-effector frame Eq. (\ref{eq:dyn_task}) includes the external torque and becomes
    \begin{equation}
        \label{eq:dyn_task_contact}
        \bm{\Lambda}\dot{\bm{\mathcal{V}}} + \bm{\eta} = \bm{\mathcal{F}} + \bar{\bm{J}}^\top \bm{\tau}_c.
    \end{equation}

    \noindent Based on Eq. (\ref{eq:dyn_task_contact}), the desired wrench $\bm{\mathcal{F}}_{ff}$ after compensating the disturbance $\bm{\tau}_c$ is designed as
    \begin{equation}
        \label{eq:wrench_ff}
        \begin{aligned}
        \bm{\mathcal{F}}_{ff} &= \bm{\Lambda} \, (\bm{K}_{p_3}(\bm{T} \ominus \bm{T}^{des}) + \bm{K}_{d_3}(\bm{\mathcal{V}}^{des} - \bm{\mathcal{V}})) \\
        &\quad + \bm{\eta} - \bar{\bm{J}}^\top \bm{\tau}_c,
        \end{aligned}
    \end{equation}    
    
    \noindent where $\bm{K}_{p_3}$ and $\bm{K}_{d_3}$ are diagonal gain matrices to guarantee stability, $\bm{T}^{des}$ and $\bm{\mathcal{V}}^{des}$ are computed by $\bm{q}^{des}$, $\dot{\bm{q}}^{des}$. 

    %  \in \mathbb{R}^{3\times n}
    %  \in \mathbb{R}^3
    Then, for the external torque, $\bm{\tau}_c = \bm{J}^\top_c \bm{f}_c$ holds, where $\bm{J}_c$ is the contact Jacobian, and $\bm{f}_c$ is the contact force. Considering that the robot may lose some DoFs due to the contact constraint, we define a reduced contact Jacobian \cite{Jorda_contact_2019} as 
    \begin{equation}
        \label{eq:reduce_Jacobian}
        \tilde{\bm{J}}_{c} = \bm{n}^\top_c \bm{J}_{c}, \quad 
    \end{equation}
    
    \noindent where $\bm{n}_c = \bar{\bm{J}}^\top_c \bm{\tau}_c / \| \bar{\bm{J}}^\top_c \bm{\tau}_c \|$ is a normalized vector representing the contact direction, $\tilde{\bm{J}}_{c}$ maps the joint velocity $\bm{q}$ to a scalar linear velocity at the contact point.
    
\begin{figure}[!t]
    \vspace{-0.3cm}
    \centering
    \subfloat[]{\includegraphics[width=.24\textwidth]{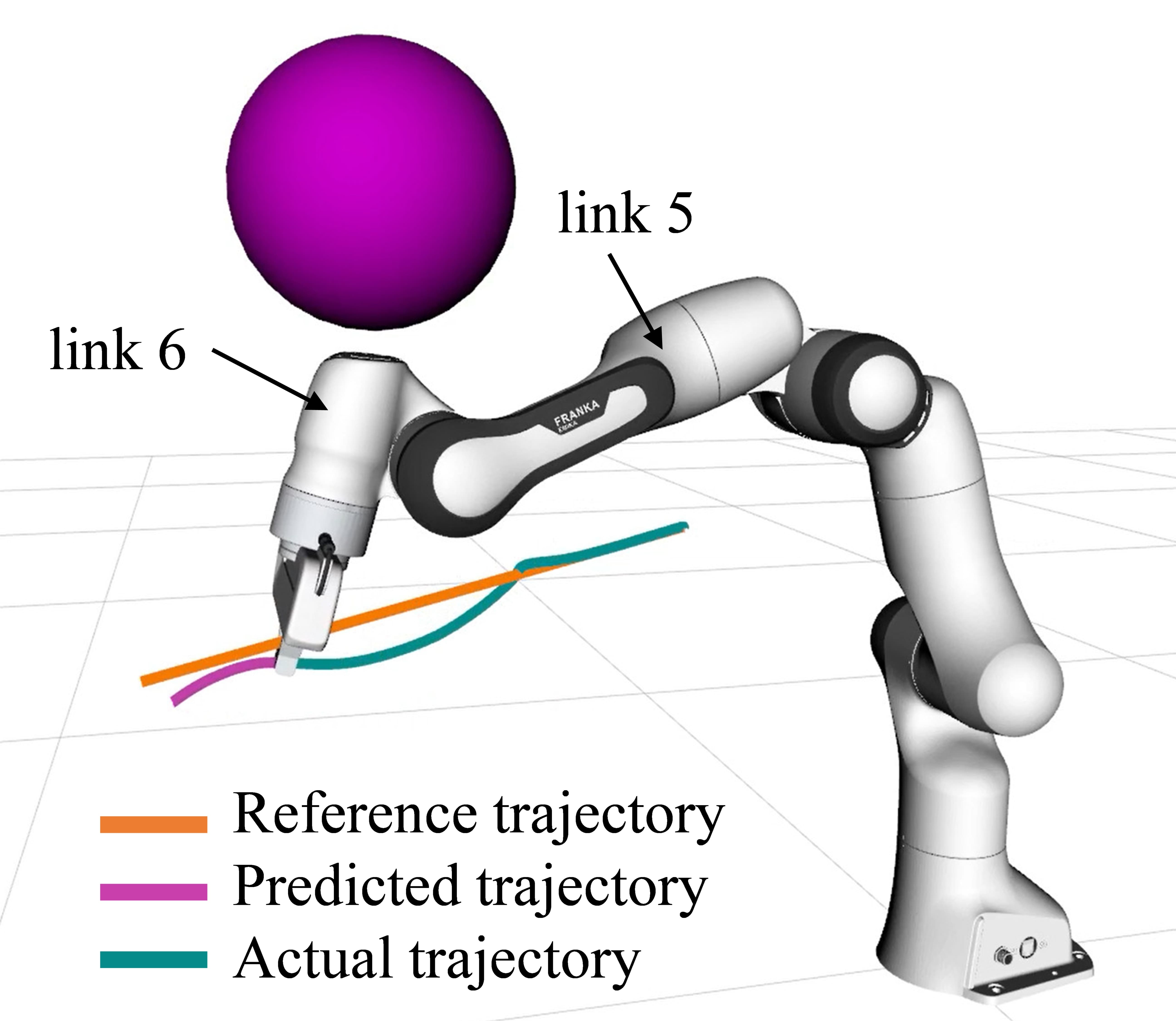}}
    \hfill
    \subfloat[]{\includegraphics[width=.24\textwidth]{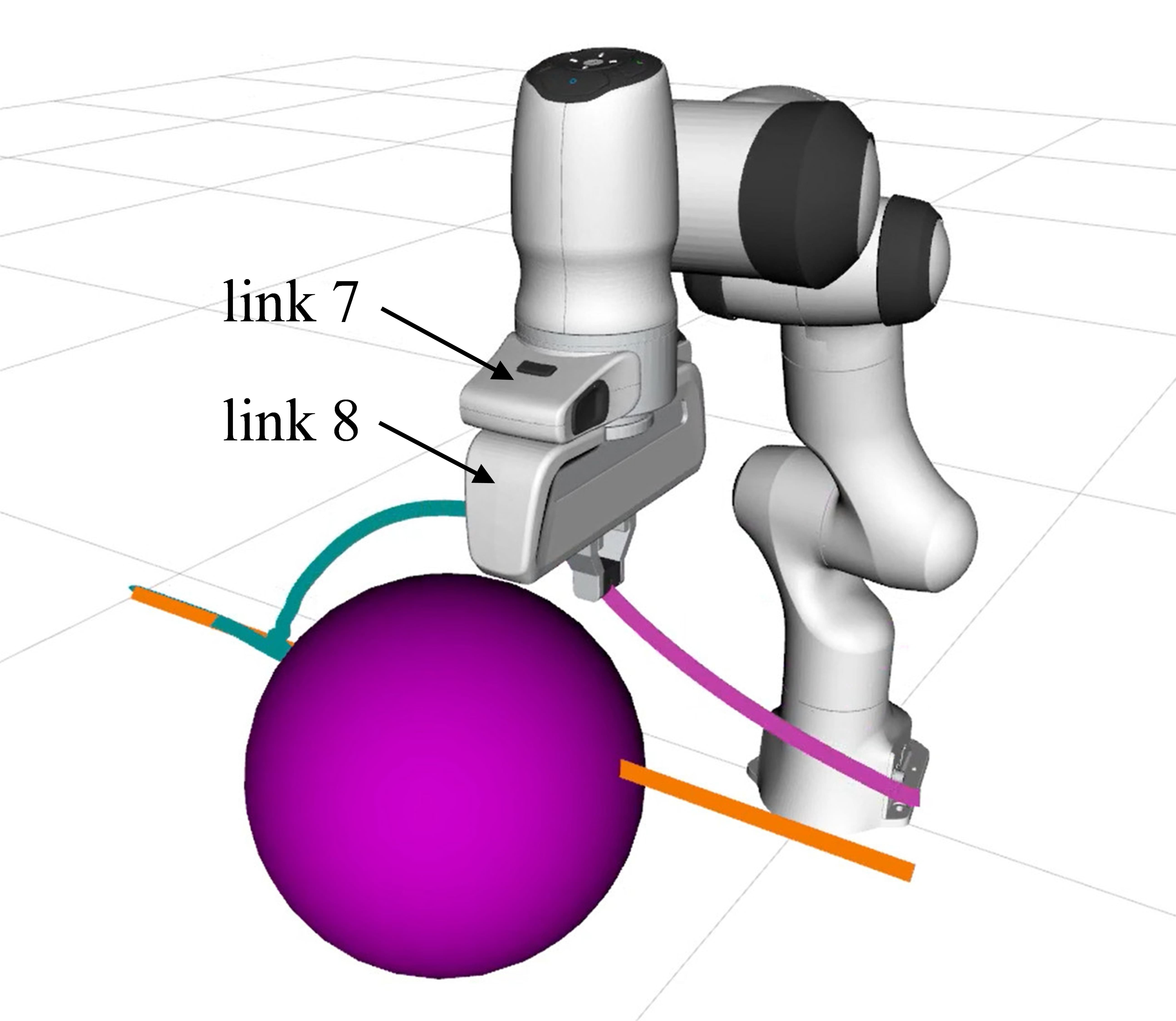}} \\
    \vspace{-0.2cm}
    \subfloat[]{\includegraphics[width=.24\textwidth]{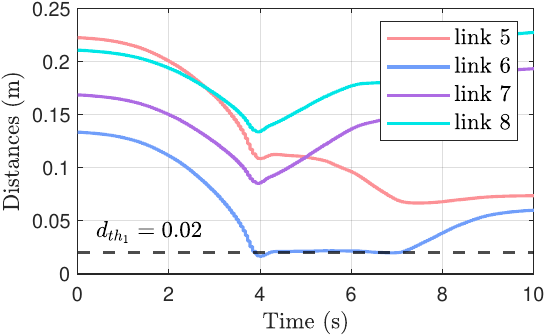}}
    \hfill
    \subfloat[]{\includegraphics[width=.24\textwidth]{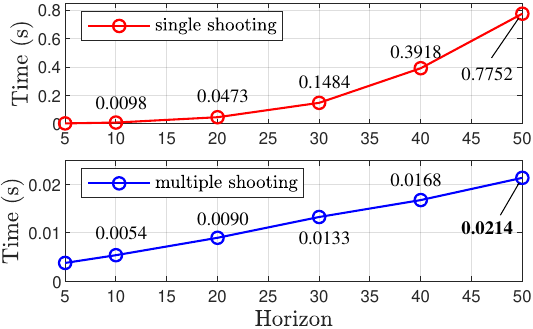}} \\
    \caption[]{Simulations of obstacle avoidance. (a) and (b) are snapshots. (c) shows the distances between the robot links and the obstacle for the top-left scenario. (d) compares solution times using different shooting methods and horizons.}
    \label{fig:sim}
    \vspace{-0.3cm}
\end{figure}

    Although the exact contact point is unknown, the involved link is known as described in Section \ref{contact_detect}. Hence, we will control the link motion for the purpose of collision reaction. The contact-safe controller is finally given by
    \begin{equation}
        \label{eq:tau_des_contact}
        \bm{\tau}^{des} = \bm{J}^\top \bm{\mathcal{F}}_{ff} + \bm{N}_t \tilde{\bm{J}}_c f^{des}
    \end{equation}
    \noindent where $\bm{N}_t$ projects the  motion of contact link into the null space of the end-effctor task, $f^{des}$ is the desired force applying in the direction of the contact constraint.

\subsection{Overview of Control Strategy}
    To achieve continuous manipulation, we set the following control strategy that could switch among the aforementioned torque controllers: (a) the robot tracks desired trajectories in free space using Eq. (\ref{eq:tau_des_free}); (b) it will switch to the contact-safe mode Eq. (\ref{eq:tau_des_contact}) upon detecting an unforeseen collision; (c) after responding to the collision, a new trajectory is generated to guide the robot back; (d) the manipulation resumes if the robot returns to its pre-contact configuration.

    % \begin{equation}
    % \label{eq:auxiliary}
    %   \begin{aligned}
    %     &\bm{\mathcal{P}}(\bm{q},\dot{\bm{q}}) = \bm{M}\dot{\bm{q}} , \\
    %     &\bm{\mathcal{H}}(\bm{q},\dot{\bm{q}}) = - \bm{C}^\top \dot{\bm{q}} + \bm{g} ,
    %   \end{aligned}
    % \end{equation}
    
    % \begin{equation}
    % \label{eq:filter}
    %   \left\{
    %   \begin{aligned}
    %     &k\dot{\bm{\mathcal{P}}}_f + \bm{\mathcal{P}}_f = \bm{\mathcal{P}}, \; & \bm{\mathcal{P}}_f|_{t=0} = \bm{0} \\
    %     &k\dot{\bm{\mathcal{H}}}_f + \bm{\mathcal{H}}_f = \bm{\mathcal{H}}, \; & \bm{\mathcal{H}}_f|_{t=0} = \bm{0} \\
    %     &k\dot{\bm{\tau}}_f + \bm{\tau}_f = \bm{\tau}, \; & \bm{\mathcal{\tau}}_f|_{t=0} = \bm{0}
    %   \end{aligned}
    %   \right.
    % \end{equation}

\section{Simulations and experiments}
\label{verification}

The proposed planning and control framework, featuring multi-layered safety, is validated in simulations and experiments on a seven-DoF Franka Emika Panda robot equipped with a two-finger gripper. The robot modeling and distance computations are handled by Pinocchio \cite{pinocchio_2019} with FCL \cite{Pan_fcl_2012} support. The MPC planner is constructed using symbolic expressions \cite{CasADi_2019} and solved by IPOPT \cite{IPOPT_2006} at a frequency of 20-100 Hz, while the torque controllers run at 1 kHz. All algorithms are implemented in C++17 on a computer with Intel i9-9900K CPU and are open-sourced to benefit the community.

\begin{figure}[!t]
    \centering
    \subfloat[Baseline]{\includegraphics[width=.24\textwidth]{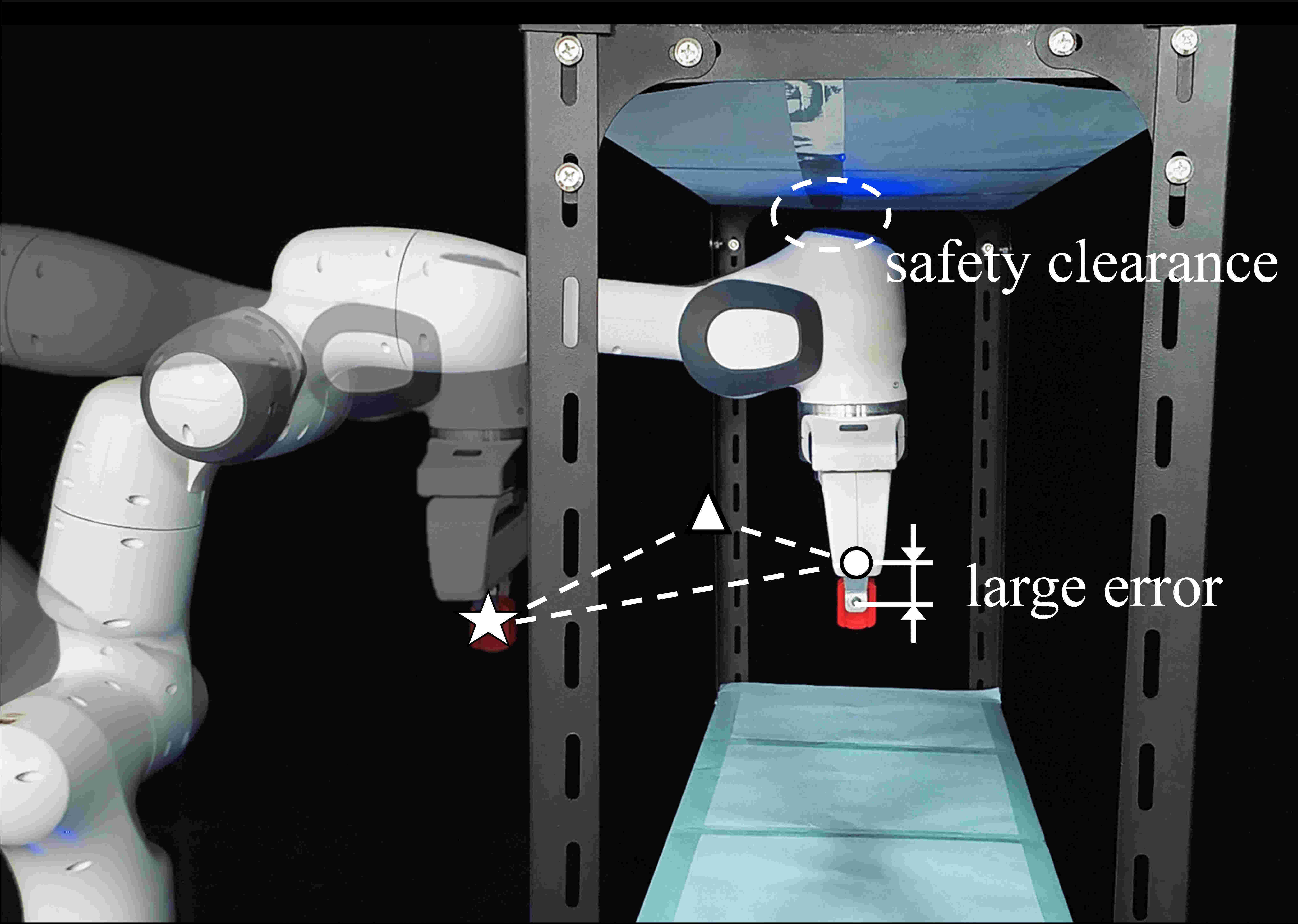}}
    \hfill
    \subfloat[Proposed]{\includegraphics[width=.24\textwidth]{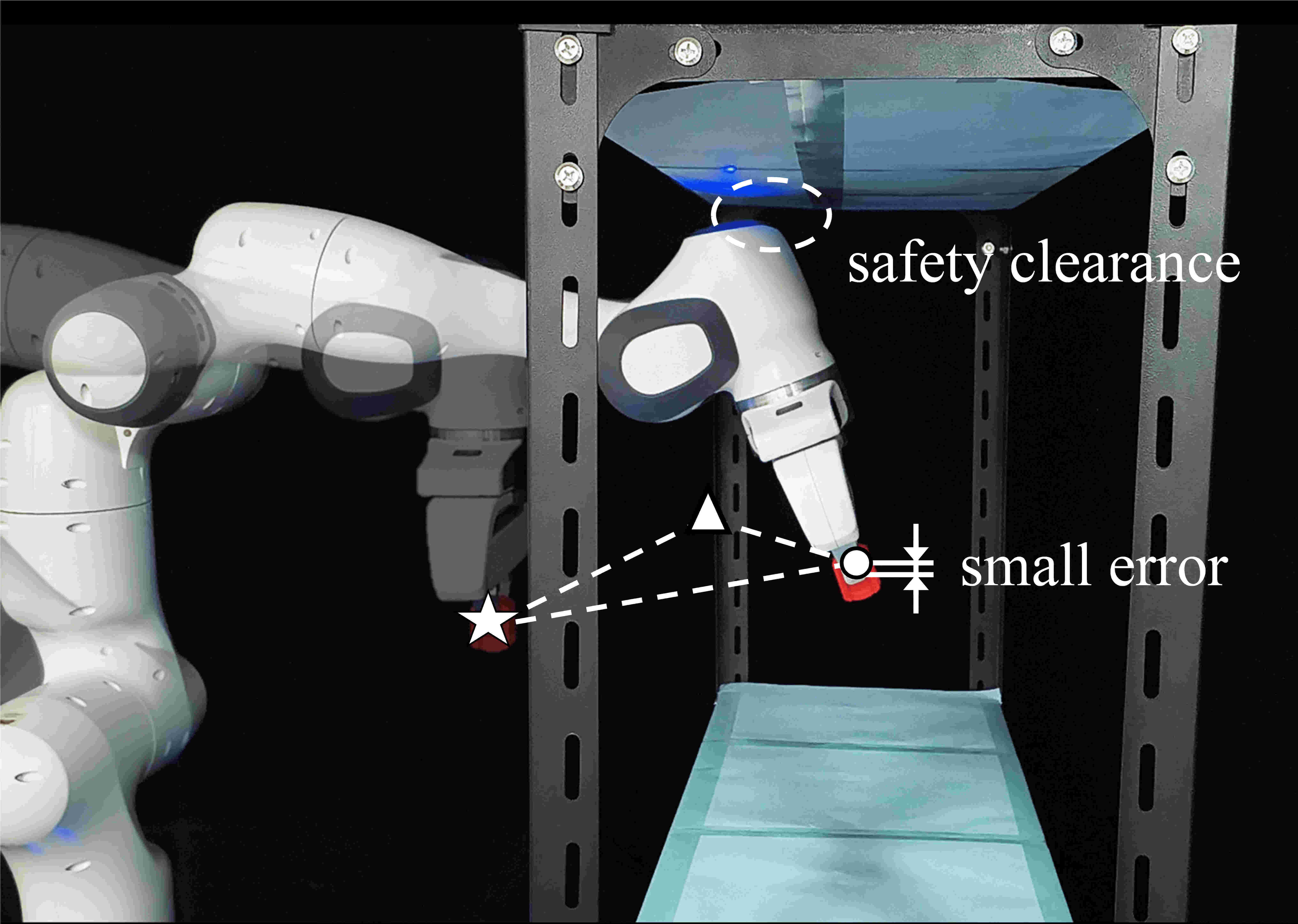}} \\
    \vspace{-0.2cm}
    \subfloat[]{\includegraphics[width=.48\textwidth]{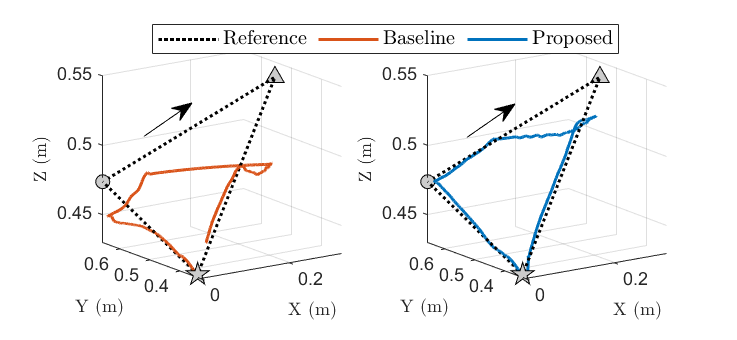}}\\
    \vspace{-0.2cm}
    \subfloat[]{\includegraphics[width=.45\textwidth]{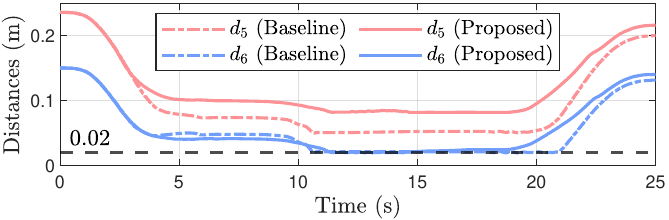}} \\
    \caption[]{Case study \#1 of \textbf{S1} and \textbf{S2} under static obstacles. (a) and (b) are experimental snapshots. (c) plots the end-effector trajectories. (d) presents the closest distances $d_i$ between the link 5-6 and the cabinet's top board.}
    \label{fig:case1}
    \vspace{-0.3cm}
\end{figure}

In terms of parameter selection, most parameters remain consistent throughout verification. The cost weights are
$\bm{Q}_{ee}$ = $\bm{I}$,
$\bm{Q}_{rep}$ = $\bm{Q}_{s}$ = $0.01\bm{I}$, 
$\bm{Q}_{ee_f}$ = $\bm{I}$,
$\bm{Q}_{s_f}$ = $10\bm{I}$, 
$\bm{R}$ = $10^{-9}\bm{I}$. 
The filter coefficient for collision detection is $k$ = $0.2$. 
The PD control gains are 
$\bm{K}_{p_1}$ = $200\bm{I}$, 
$\bm{K}_{d_1}$ = $10\bm{I}$,
$\bm{K}_{p_2}$ = $10\bm{I}$, 
$\bm{K}_{d_2}$ = $2\bm{I}$,
$\bm{K}_{p_3}$ = $500\bm{I}$, 
$\bm{K}_{d_3}$ = $100\bm{I}$.
In real-world experiments, the joint-space gains are modified as $\bm{K}_{p_2}$ = $50\bm{I}$, $\bm{K}_{d_2}$ = $5\bm{I}$ to compensate for model errors. The three thresholds defining safety regions are $d_{th_1}$ = 0.02 m, $d_{th_2}$ = 0.1 m, and $\tau_{th}$ = 3 N.m. In the following, we evaluate the MPC in simulation and validate the multi-layered safety in three real-hardware cases.

\subsection{Evaluation of MPC Performance}
Two simulation scenarios are set in ROS/Gazebo to evaluate the basic capability (i.e. only \textbf{S2}) of the proposed MPC. As shown in Fig. \ref{fig:sim} (a) and (b), the robot generates smooth, collision-free trajectories in the presence of a spherical obstacle. Fig. \ref{fig:sim} (c) records the distances of robot links to the overhead sphere, with all distances kept beyond a safety threshold of 0.02 m, thereby ensuring manipulation safety. We also compare solution times using different shooting methods and horizons $N$ with $\Delta t$ = 0.05 s, see Fig. \ref{fig:sim} (d). The results show that the multiple shooting can accelerate computation, particularly for long horizons. In this paper, we select multiple shooting and $N$ = 50 for all cases, corresponding to a time horizon of 2.5 s.

\begin{figure}[!t]
    \centering
    \subfloat[Baseline]{\includegraphics[width=.24\textwidth]{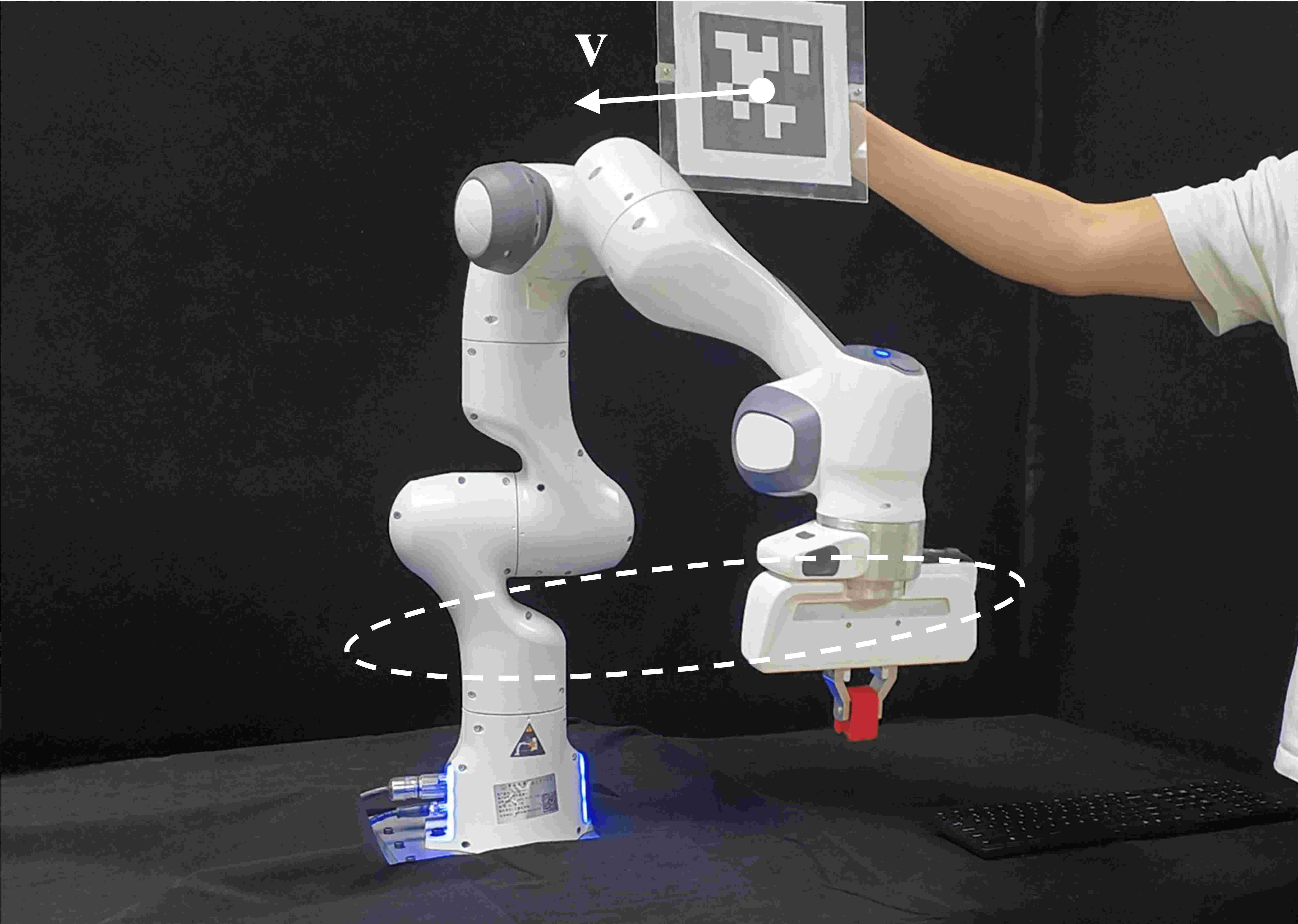}}
    \hfill
    \subfloat[Proposed]{\includegraphics[width=.24\textwidth]{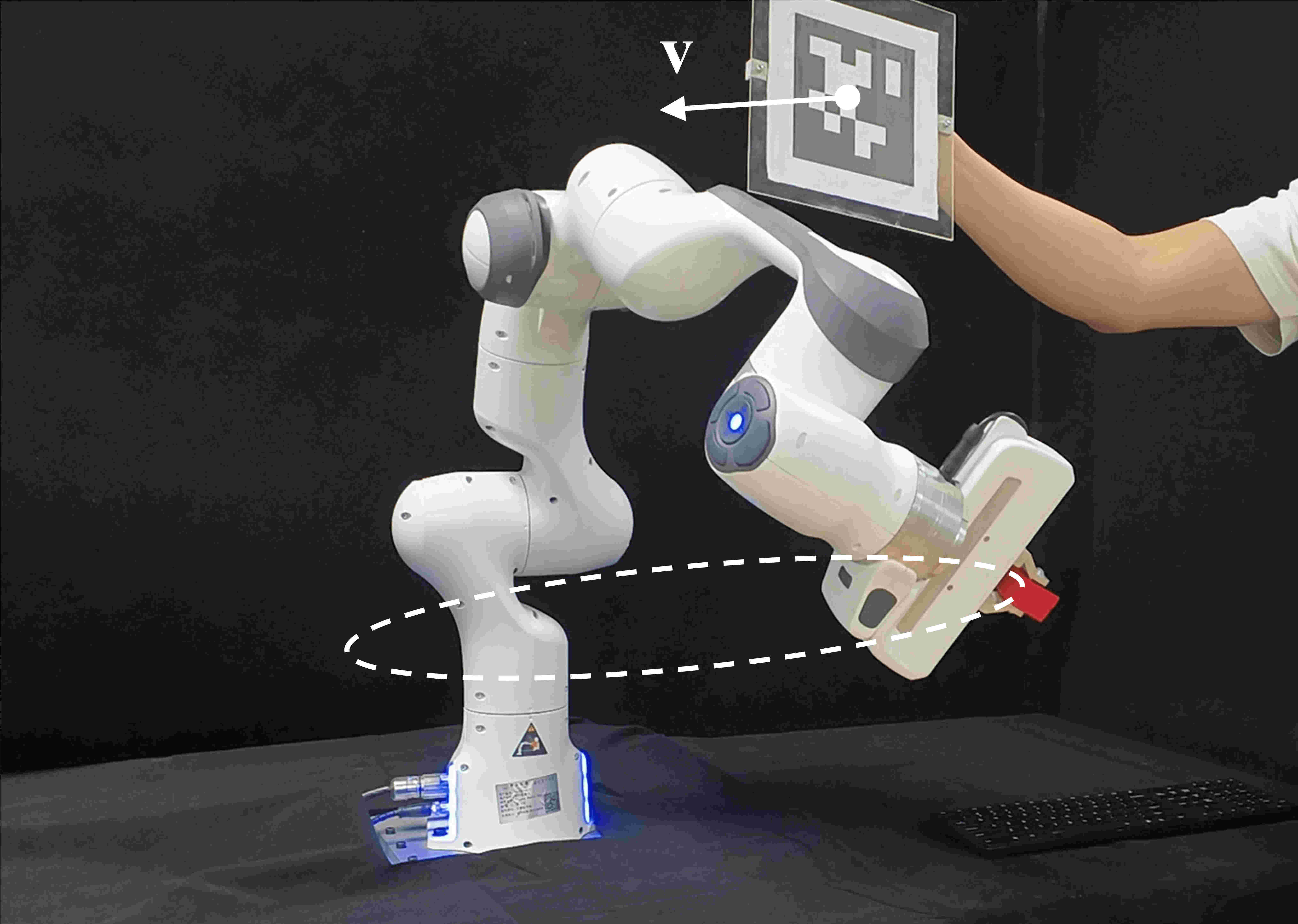}}\\
    \vspace{-0.2cm}
    \subfloat[]{\includegraphics[width=.48\textwidth]{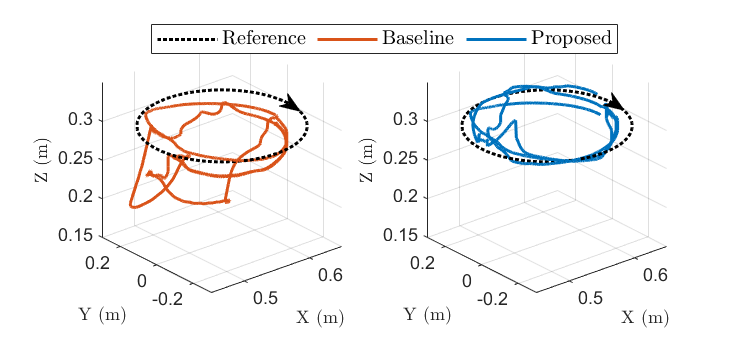}}\\
    \vspace{-0.2cm}
    \subfloat[]{\includegraphics[width=.45\textwidth]{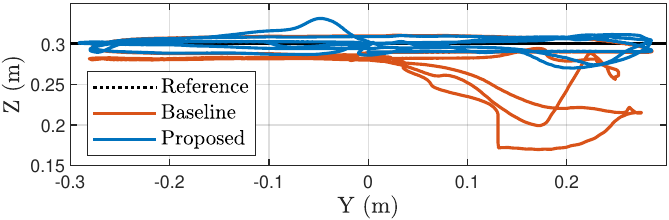}}\\
    \caption[]{Case study \#2 of \textbf{S1} and \textbf{S2} under dynamic obstacles. (a) and (b) are experimental snapshots. (c) and (d) plot the end-effector trajectories.}
    \label{fig:case2}
    \vspace{-0.3cm}
\end{figure}

\begin{figure*}[!ht]
    \centering
    \subfloat[]{\includegraphics[width=.193\textwidth]{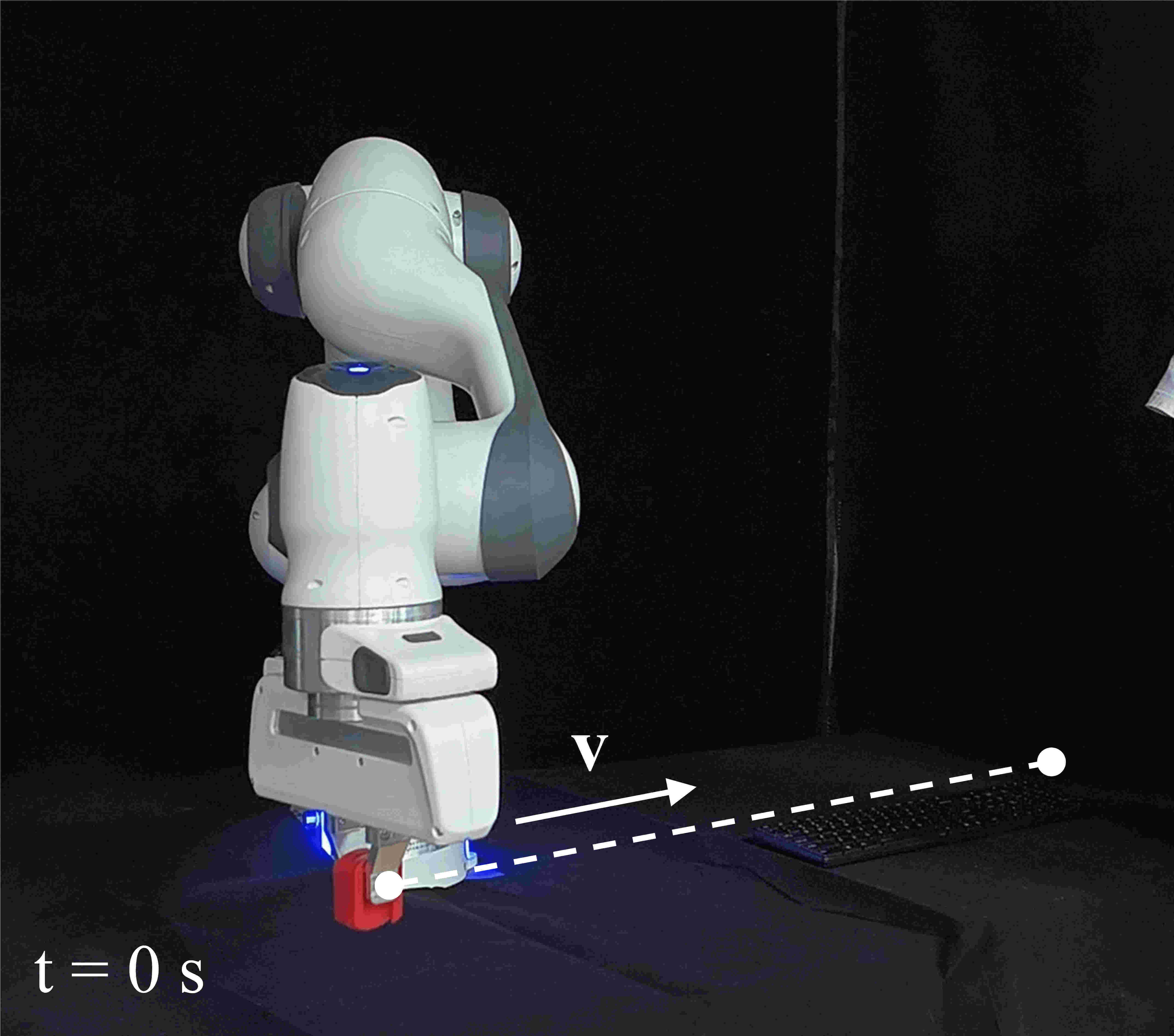}}
    \hfill
    \subfloat[]{\includegraphics[width=.193\textwidth]{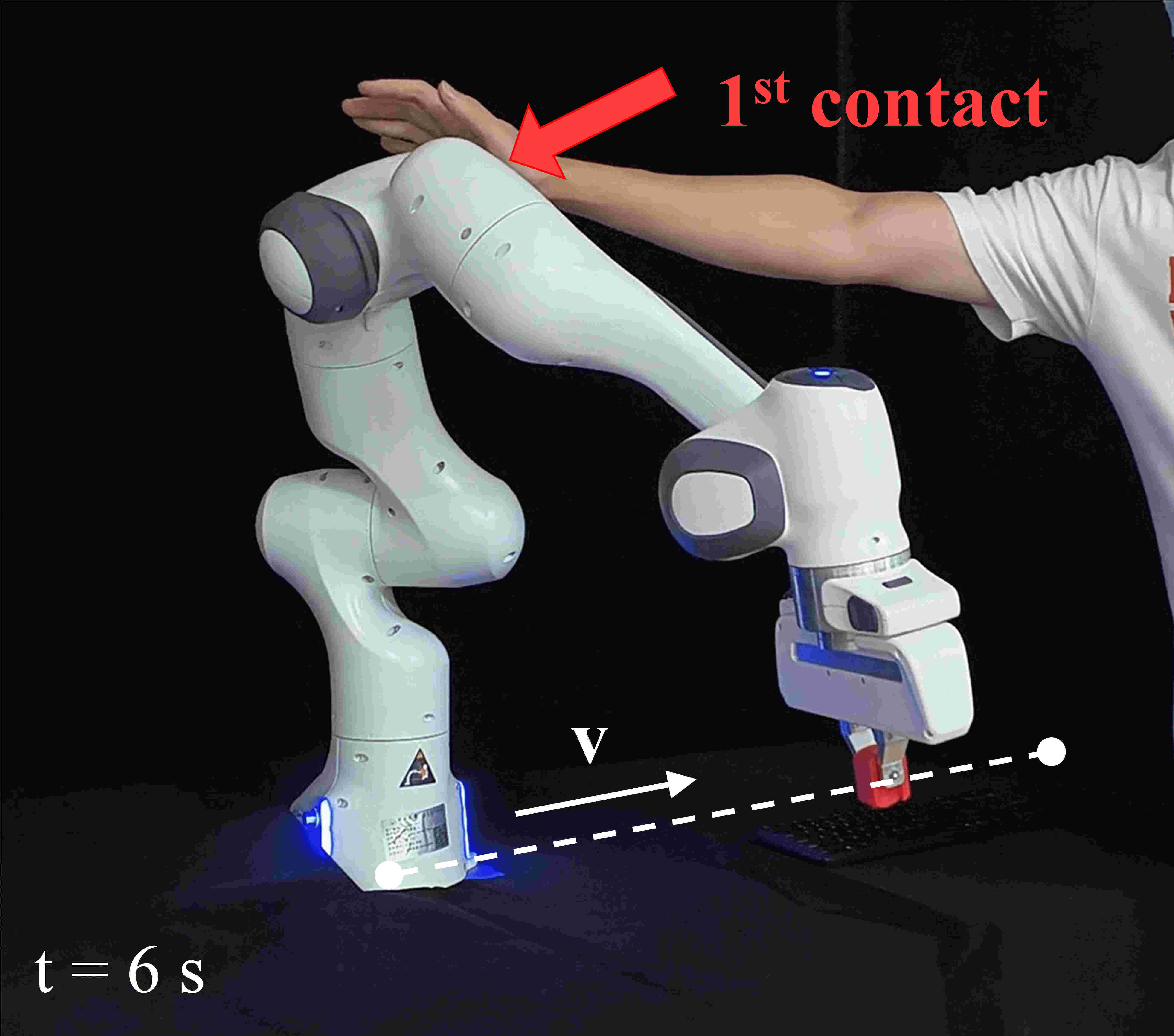}}
    \hfill
    \subfloat[]{\includegraphics[width=.193\textwidth]{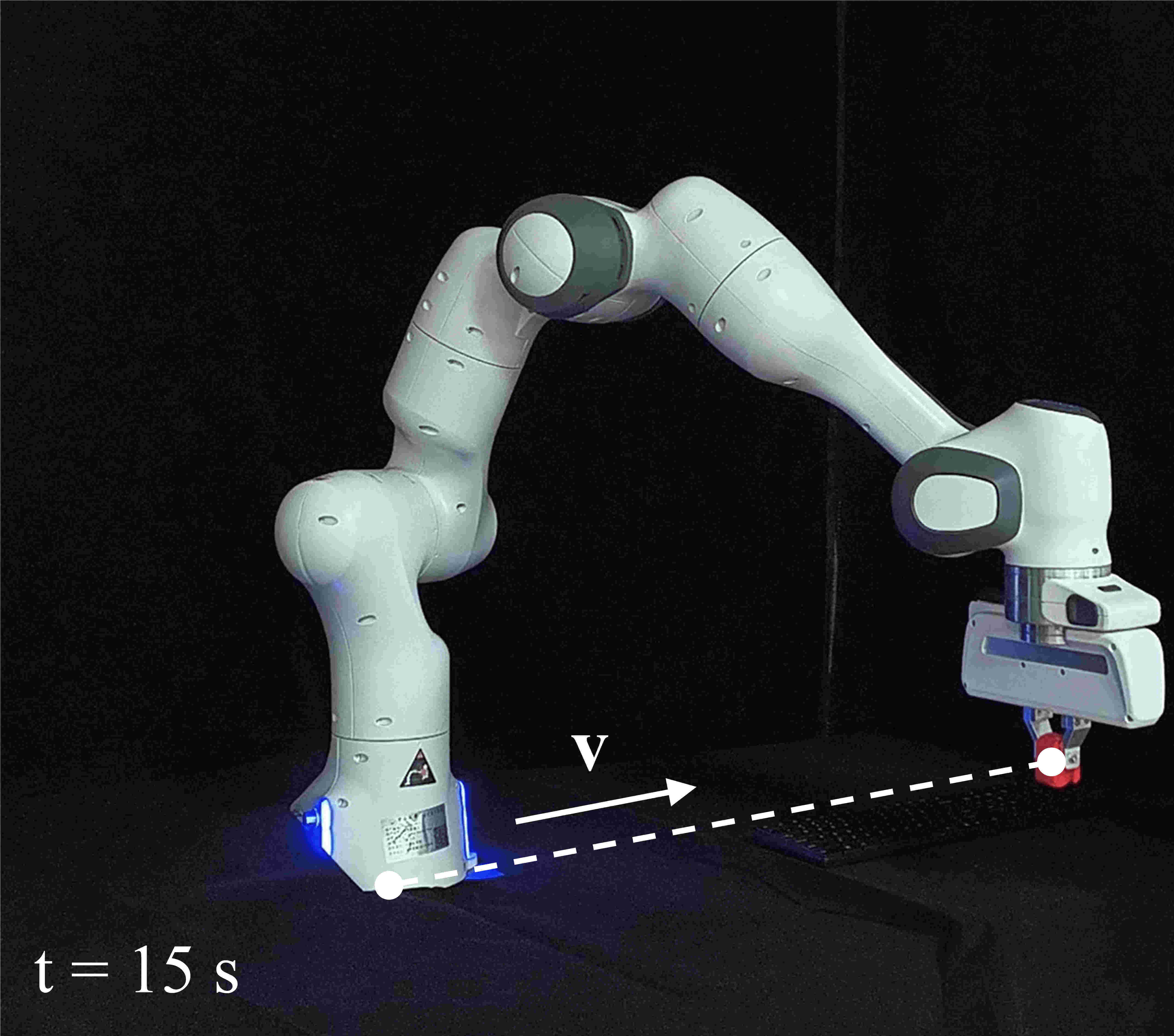}}
    \hfill
    \subfloat[]{\includegraphics[width=.193\textwidth]{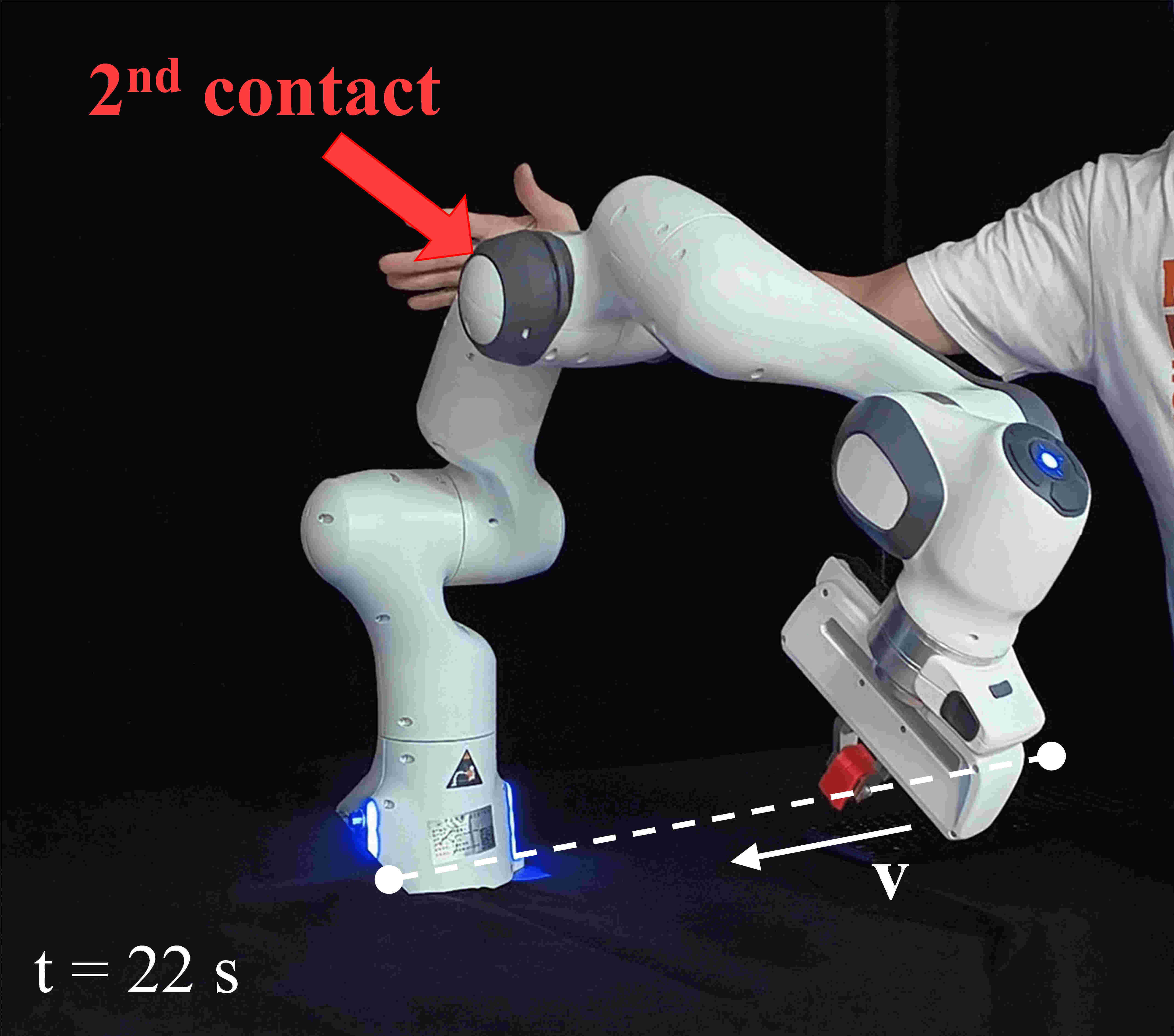}}
    \hfill
    \subfloat[]{\includegraphics[width=.193\textwidth]{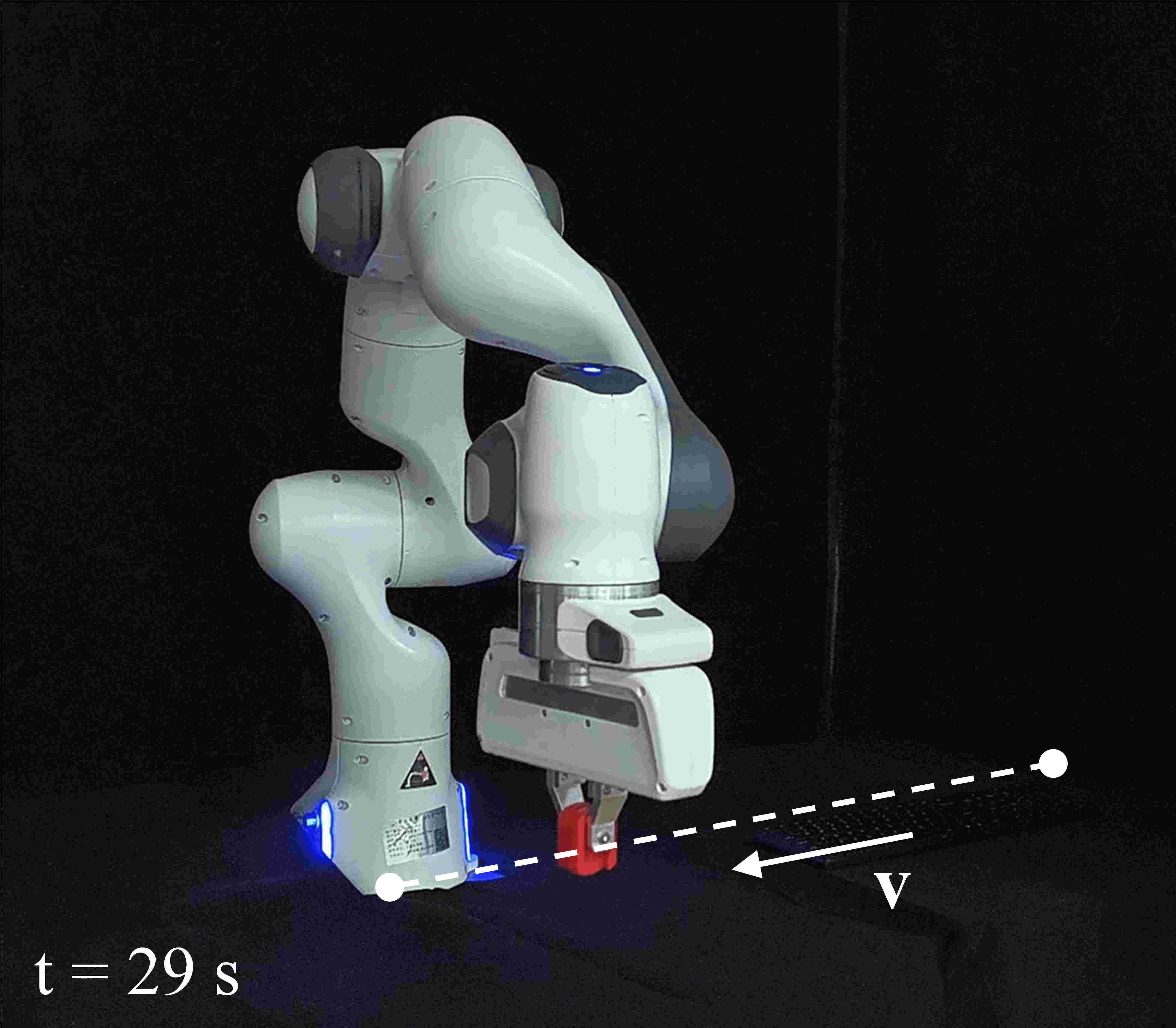}} \\
    \caption[]{Case study \#3 of \textbf{S3} under unknown contacts. (a)-(e) are experimental snapshots, where (b) and (d) show two collisions caused by human pushes.}
    \label{fig:case3}
    \vspace{-0.2cm}
\end{figure*}

\subsection{Case Study \#1 - Static Obstacles}
\label{case1}
The first experiment is conducted to validate the capability of the proposed scheme in ensuring  \textbf{S1} and \textbf{S2}. As illustrated in Fig. \ref{fig:case1}, the robot manipulates a red block into a cabinet (i.e., static obstacle). For simplicity, the reference trajectory is obtained by interpolating between three target points. A baseline which only considers \textbf{S2} is compared. The experimental snapshots show that the two methods produce different avoidance motions when the links approach the top board. The end-effector trajectories are plotted in Fig. \ref{fig:case1} (c). Compared to the baseline, the proposed task-oriented avoidance approach results in smaller errors at every target point. Additionally, Fig. \ref{fig:case1} (d) depicts the closest distances $d_i$ between the link 5-6 and the obstacle, where the planner considering \textbf{S1} provides more safety clearance, reducing potential collision risk due to environmental uncertainties (e.g., inaccurate obstacle position).

\begin{figure}[!t]
    \centering
    \subfloat[]{\includegraphics[width=.46\textwidth]{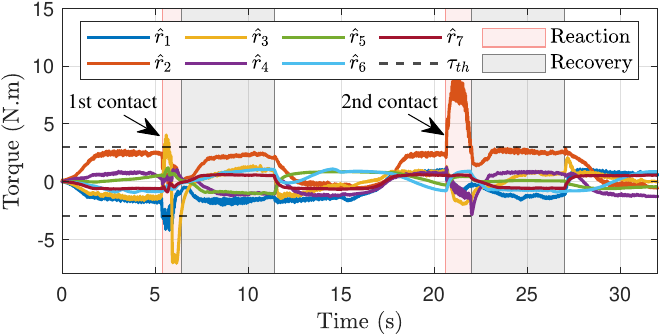}}\\
    \vspace{-0.2cm}
    \subfloat[]{\includegraphics[width=.46\textwidth]{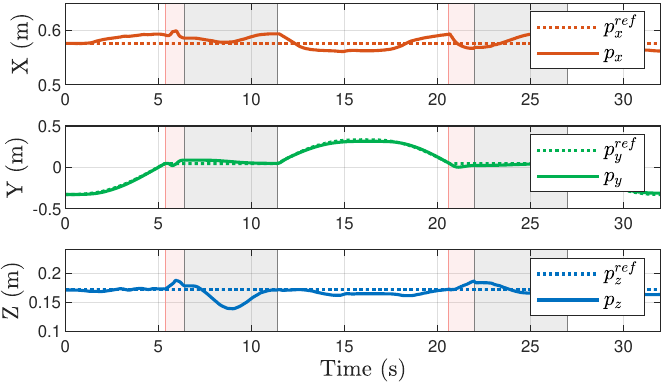}}\\
    \caption[]{Experimental results in case study \#3. (a) shows the estimation of external torques. (b) presents the desired and actual end-effector positions.}
    \label{fig:case3data}
    \vspace{-0.3cm}
\end{figure}

\subsection{Case Study \#2 - Dynamic Obstacles}
\label{case2}
To further demonstrate the superiority of the proposed framework, we conduct an experiment involving a dynamic obstacle. As described in Fig. \ref{fig:case2}, while the robot follows a circular trajectory, a marker (i.e., the obstacle) is introduced into the robot workspace and detected by a RealSense D435i RGB-D camera. The position of this nonstationary obstacle relative to the robot base is reported by ViSP \cite{ViSP_2005} at 30 Hz. The baseline still neglects \textbf{S1}. Notably, despite the different obstacle locations, the proposed method shows superior performance over the baseline, which allows the robot to twist its whole body to optimize task execution. In contrast, the baseline exhibits larger tracking errors in terms of the task space position, though robot safety is guaranteed, see Fig. \ref{fig:case2} (c) and (d).

\subsection{Case Study \#3 - Unknown Contacts}
\label{case3}
In the third experiment, we take into account the effect of unforeseen collisions in robot manipulation (i.e., \textbf{S3}). The end-effector is commanded to move along a straight line, and two intentional contacts are successively applied to the robot links. Fig. \ref{fig:case3} presents a photo series taken at different time points during the task execution. Expectedly, the robot complies with the contact by moving in the direction of the human push. Fig. \ref{fig:case3data} (a) shows the estimated external torques $\hat{\bm{r}}$ for all seven joints. It can be seen that the contacts at 6 and 22 seconds are effectively detected. In contrast to the baseline which directly shuts down motors for protection (see the video attachment), the proposed method demonstrates safer behavior by reducing the external torque as the robot actively moves away from the collision area. Moreover, as illustrated in Fig. \ref{fig:case3data} (b), the manipulation task is minimally affected, which is indicated by the small end-effector position errors. In other words, the task consistency is maintained in collisions.

% error, filter delay

\section{Conclusion}
\label{conclusion}

In this paper, we propose a task-oriented planning and control framework for redundant robot manipulators to achieve multi-layered safety in uncertain environments. Therein, a novel trajectory planner based on nonlinear MPC is formulated to minimize collision risks in obstacle avoidance, and a contact-safe torque controller is designed to handle unexpected collisions or contacts using proprioceptive sensors. In addition to safety guarantee during manipulation, the framework also emphasizes maximizing task execution efficiency. Simulations and experiments on a seven-DoF manipulator demonstrate the superior performance of proposed method compared to the baselines. Although our work has shown promising results, there remains room for improvement. For instance, simply using a fixed threshold $\tau_{th}$ to detect collisions may lead to false alarms, highlighting the need for a flexible decision mechanism. Besides, in future work, we plan to integrate richer perceptive information for obstacle avoidance, and allocate additional task layers to better exploit the manipulator's redundancy.

% \section*{Appendix}
% \input{sections/appendix}

% \section*{Acknowledgment}
% The authors are grateful to Terry Cavan Chan for their help in experiments. 

\addtolength{\textheight}{-7.5cm}

\bibliographystyle{IEEEtran}
\bibliography{references}

\end{document}